# RUSSE: THE FIRST WORKSHOP ON RUSSIAN SEMANTIC SIMILARITY

**Panchenko A.** (panchenko@lt.informatik.tu-darmstadt.de)
TU Darmstadt, Darmstadt, Germany
Université catholique de Louvain, Louvain-la-Neuve, Belgium

**Loukachevitch N. V.** (louk_nat@mail.ru)
Moscow State University, Moscow, Russia

**Ustalov D.** (dau@imm.uran.ru)
N. N. Krasovskii Institute of Mathematics and Mechanics,
Ural Branch of the RAS, Russia;
NLPub, Yekaterinburg, Russia

**Paperno D.** (denis.paperno@unitn.it)
University of Trento, Rovereto, Italy

**Meyer C. M.** (meyer@ukp.informatik.tu-darmstadt.de)
TU Darmstadt, Darmstadt, Germany

**Konstantinova N.** (n.konstantinova@wlv.ac.uk)
University of Wolverhampton, Wolverhampton, UK

The paper gives an overview of the Russian Semantic Similarity Evaluation (RUSSE) shared task held in conjunction with the Dialogue 2015 conference. There exist a lot of comparative studies on semantic similarity, yet no analysis of such measures was ever performed for the Russian language. Exploring this problem for the Russian language is even more interesting, because this language has features, such as rich morphology and free word order, which make it significantly different from English, German, and other well-studied languages. We attempt to bridge this gap by proposing a shared task on the semantic similarity of Russian nouns. Our key contribution is an evaluation methodology based on four novel benchmark datasets for the Russian language. Our analysis of the 105 submissions from 19 teams reveals that successful approaches for English, such as distributional and skip-gram models, are directly applicable to Russian as well. On the one hand, the best results in the contest were obtained by sophisticated supervised models that combine evidence from different sources. On the other hand, completely unsupervised approaches, such as a skip-gram model estimated on a large-scale corpus, were able score among the top 5 systems.

## 1. Introduction

A *similarity measure* is a numerical measure of the degree two given objects are alike. A *semantic similarity measure* is a specific kind of similarity measure designed to quantify the similarity of two lexical items such as nouns or multiword expressions. It yields high values for pairs of words in a semantic relation (synonyms, hyponyms, free associations, etc.) and low values for all other, unrelated pairs.

Semantic similarity measures proved useful in text processing applications, including text similarity, query expansion, question answering and word sense disambiguation [28]. A wide variety of measures were proposed and tested during the last 20 years, ranging from lexical-resource-based [31] to vector-based approaches, which in their turn evolved from Hyperspace Analogue to Language (HAL) by Lund and Burgess [24] to Latent Semantic Analysis (LSA) by Landauer and Dumais [20], topic models [12], Distributional Memory [2] and finally to neural network language models [26]. Many authors tried to perform exhaustive comparisons of existing approaches and developed a whole range of benchmarks and evaluation datasets. See Lee [22], Agirre et al. [1], Ferret [8], Panchenko [28], Baroni [4], Sahlgren [33], Curran [7], Zesch and Gurevych [38] and Van de Cruys [36] for an overview of the state-of-the-art techniques for English. A recent study of semantic similarity for morphologically rich languages, such as German and Greek, by Zervanou et al. [40] is relevant to our research. However, Russian is not considered in the latter experiment.

Unfortunately, most of the approaches to semantic similarity were implemented and evaluated only on a handful of European languages, mostly English. Some researchers, such as Krizhanovski [18], Turdakov [35], Krukov et al. [19] and Sokirko [34], worked towards adapting several methods developed for English to the Russian language. These efforts were, however, mostly done in the context of a few specific applications without a systematic evaluation and model comparison. To the best of our knowledge, no systematic investigation of semantic similarity measures for Russian was ever performed.

The very goal of the *Russian Semantic Similarity Evaluation* (*RUSSE*) shared task[1] is to fill this gap, conducting a systematic comparison and evaluation of semantic similarity measures for the Russian language. The event is organized as a competition where systems are calculating similarity between words of a joint, previously unseen gold standard dataset.

To this end, we release four novel test datasets for Russian and an open-source tool for evaluating semantic similarity measures[2]. Using this standardized evaluation methodology, we expect that each new semantic similarity measure for the Russian language can be seamlessly compared to the existing ones. To the best of our knowledge, *RUSSE* is the largest and most comprehensive evaluation of Russian similarity measures to date.

This paper is organized as follows: First, we describe previous shared tasks covering other languages. In Section 3, we outline the proposed evaluation methodology. Finally, Section 4 presents the key results of the shared task along with a brief discussion.

---

[1]  http://russe.nlpub.ru

[2]  https://github.com/nlpub/russe-evaluation/tree/master/russe/evaluation

## 2. Related Work

Evaluation of semantic similarity approaches can be fulfilled in various settings [3, 6, 21]. We identified three major research directions which are most related to our shared task.

**The first strand of research** is testing of automatic approaches relative to human judgments of word pair similarity. Most known gold standards for this task include the *RG* dataset [32], the *MC* dataset [27] and *WordSim353* [9]. These datasets were created for English. To enable similar experiments in other languages, there have been several attempts to translate these datasets into other languages. Gurevych translated the *RG* and *MC* datasets into German [13]; Hassan and Mihalcea translated them into Spanish, Arabic and Romanian [14]; Postma and Vossen [29] translate the datasets into Dutch; Jin and Wu [15] present a shared task for Chinese semantic similarity, where the authors translated the *WordSim353* dataset. Yang and Powers [37] proposed a dataset specifically for measuring verb similarity, which was later translated into German by Meyer and Gurevych [25].

Hassan and Mihalcea [14] and Postma and Vossen [29] divide their translation procedure into the following steps: disambiguation of the English word forms; selection of a translation for each word; additionally, translations were checked to be in the same relative frequency class as the source English word.

**The second strand of research** consists in testing of automated systems with respect to relations described in a lexical-semantic resource such as WordNet. Baroni and Lenci [3] stress that semantically related words differ in the type of relations between them, so they generate the BLESS dataset containing tuples of the form (w1, w2, relation). Types of relations include COORD (co-hyponyms), HYPER (hypernyms), MERO (meronyms), ATTRI (attributes—relation between a noun and an adjective expressing an attribute), EVENT (relation between a noun and a verb referring to actions or events). BLESS also contains, for each concept, a number of random words that were checked to be semantically unrelated to the target word. BLESS includes 200 English concrete single-word nouns having reasonably high frequency that are not very polysemous. The relata of the non-random relations are English nouns, verbs and adjectives selected and validated using several sources including WordNet, Wikipedia and the Web-derived ukWaC corpus.

**The third strand of research** evaluates possibilities of current automated systems to simulate the results of human word association experiments. The task originally captured the attention of psychologists, such as Griffiths and Steyvers [10–11]. One such task was organized in the framework of the CogALex workshop [30]. The participants received lists of five given words (primes) such as *circus*, *funny*, *nose*, *fool*, and *Coco* and were supposed to compute the word most closely associated to all of them. In this specific case, the word *clown* would be the expected response. 2,000 sets of five input words, together with the expected target words (associative responses) were provided as a training set to the participants. The test dataset contained another 2,000 sets of five input words. The training and the test datasets were both derived from the *Edinburgh Associative Thesaurus* (EAT) [16]. For each stimulus word, only the top five associations, i.e. the associations produced by the largest number of respondents, were retained, and all other associations were discarded.

## 3. Evaluation Methodology

In this section, we describe our approach to the evaluation of Russian semantic similarity measures used in the *RUSSE* shared task. Each participant had to calculate similarities between 14,836 word pairs[3]. Each submission was assessed on the following four benchmarks, each being a subset of these 14,836 word pairs:

1. **HJ.** Correlations with human judgments in terms of Spearman's rank correlation. This test set was composed of 333 word pairs.
2. **RT.** Quality of semantic relation classification in terms of average precision. This test set was composed of 9,548 word pairs (4,774 unrelated pairs and 4,774 synonyms and hypernyms from the *RuThes-lite* thesaurus[4]).
3. **AE**. Quality of semantic relation classification in terms of average precision. This test set was composed of 1,952 word pairs (976 unrelated pairs and 976 cognitive associations from the *Russian Associative Thesaurus*[5]).
4. **AE2.** Quality of semantic relations classification in terms of average precision. This test set was composed of 3,002 word pairs (1,501 unrelated pairs and 1,501 cognitive associations from a large-scale web-based associative experiment[6]).

In order to help participants to build their systems, we provided training data for each of the benchmarks (see Table 1). In case of the *HJ* dataset, it was only a small validation set of 66 pairs as annotation of word pairs is expensive. On the other hand, for the *RT*, *AE* and *AE2,* we had prepared substantial training collections of 104,518, 20,968, and 104,518 word pairs, respectively.

We did not limit the number of submissions per participant. Therefore, it was possible to present several models each optimised for a given type of semantic relation: synonyms, hypernyms or free associations. We describe each benchmark dataset below and summarize their key characteristics in Table 1.

**Table 1.** Evaluation datasets used in the RUSSE shared task

| Name | Description | Source | #word pairs, test | #word pairs, train |
|---|---|---|---|---|
| HJ | human judgements | Crowdsourcing | 333 | 66 |
| RT | synonyms, hypernyms, hyponyms | RuThes Lite | 9,548 | 104,518 |
| AE | cognitive associations | Russian Associative Thesaurus | 1,952 | 20,968 |
| AE2 | cognitive associations | Sociation.org | 3,002 | 83,770 |

---

[3]   https://github.com/nlpub/russe-evaluation/blob/master/russe/evaluation/test.csv

[4]   http://www.labinform.ru/pub/ruthes/index.htm

[5]   http://it-claim.ru/asis

[6]   http://sociation.org

### 3.1. Evaluation based on Correlations with Human Judgments (HJ)

The first dataset is based on human judgments about semantic similarity. This is arguably the most common way to assess a semantic similarity measure. The *HJ* dataset contains word pairs translated from the widely used benchmarks for English: *MC* [27], *RG* [32] and *WordSim353* [9]. We translated all English words as Russian nouns, trying to keep constant the Russian translation of each individual English word. It is not possible to keep exact translations for all pairs that have an exact match between lexical semantic relations between the two languages because of the different structure of polysemy in English and Russian. For example, the pair *train* vs. *car* was translated as *поезд—машина* rather than *поезд—вагон* to keep the Russian equivalent of car consistent with other pairs in the datset. Evaluation metric in this benchmark is Spearman's rank correlation coefficient (ρ) between a vector of human judgments and the similarity scores. Table 2 shows an example of some relations from the *HJ* collection.

**Table 2.** Example of human judgements about semantic similarity (HJ)

| word1 | word2 | sim |
|---|---|---|
| петух (cock) | петушок (cockerel) | 0.952 |
| побережье (coast) | берег (shore) | 0.905 |
| тип (type) | вид (kind) | 0.852 |
| миля (mile) | километр (kilometre) | 0.792 |
| чашка (cup) | посуда (tableware) | 0.762 |
| птица (bird) | петух (cock) | 0.714 |
| война (war) | войска (troops) | 0.667 |
| улица (street) | квартал (block) | 0.667 |
| … | … | … |
| доброволец (volunteer) | девиз (motto) | 0.091 |
| аккорд (chord) | улыбка (smile) | 0.088 |
| энергия (energy) | кризис (crisis) | 0.083 |
| бедствие (disaster) | площадь (area) | 0.048 |
| производство (production) | экипаж (crew) | 0.048 |
| мальчик (boy) | мудрец (sage) | 0.042 |
| прибыль (profit) | предупреждение (warning) | 0.042 |
| напиток (drink) | машина (car) | 0.000 |
| сахар (sugar) | подход (approach) | 0.000 |
| лес (forest) | погост (graveyard) | 0.000 |
| практика (practice) | учреждение (institution) | 0.000 |

In order to collect human judgements, we utilized a simple crowdsourcing scheme that is similar to HITs in *Amazon Mechanical Turk*[7]. We decided to use a lightweight crowdsourcing software developed in-house due to the lack of native Russian speakers on popular platforms including *Amazon Mechanical Turk* and *CrowdFlower*[8]. The crowdsourcing process ran for 27 days from October 23 till November 19, 2014.

---

[7] https://www.mturk.com

Firstly, we set up a special section on the *RUSSE* website and asked volunteers on Facebook and Twitter to participate in the experiment. Each annotator received an assignment consisting of 15 word pairs randomly selected from the 398 preliminarily prepared pairs, and has been asked to assess the similarity of each pair. The possible values of similarity were 0—not similar at all, 1—weak similarity, 2—moderate similarity, and 3—high similarity. Before the annotators began their work, we provided them with simple instructions[9] explaining the procedure and goals of the study.

Secondly, we defined two assignment generation modes for the word pairs: 1) a pair is annotated with a probability inversely proportional to the number of current annotations (*COUNT*); 2) a pair is annotated with a probability proportional to the standard deviation of annotations (*SD*). Initially, the *COUNT* mode has been used, but during the annotation process, we changed to mode to *SD* several times.

By the end of the experiment, we obtained a total of 4,200 answers, i.e. 280 submissions of 15 judgements. Some users participated in the study twice or more, annotating a different set of pairs each time. We used Krippendorff's alpha [17] with an ordinal distance function to measure the inter-rater agreement: $\alpha = 0.49$, which is a moderate agreement. The average standard deviation of answers by pair is $\bar{\sigma} = 0.62$ on the scale 0–3. This result can be explained primarily by two facts: (1) the participants were probably confusing "weak" and "moderate" similarity, and (2) some pairs were ambiguous or too abstract. For instance, it proved difficult for participants to estimate the similarity between the words «деньги» ("money") and «отмывание» ("laundering"), because on the one hand, these words are associated, being closely connected within the concept of money laundering, while on the other hand these words are ontologically dissimilar and are indeed unrelated outside the particular context of money laundering.

### 3.2. Semantic Relation Classification of Synonyms and Hypernyms (RT)

This benchmark quantifies how well a system is able to detect synonyms and hypernyms, such as:

- автомобиль, машина, syn      (car, automobile, syn)
- кошка, животное, hypo       (cat, animal, hypo)

The evaluation dataset follows the structure of the *BLESS* dataset [3]. Each target word has the same number of related and unrelated source words as exemplified in Table 3. First, we gathered 4,774 synonyms and hypernyms from the *RuThes Lite* thesaurus [23]. We used only single word nouns at this step. These relations were considered positive examples. To generate negative examples we used the following procedure:

---

[8]   http://www.crowdflower.com

[9]   http://russe.nlpub.ru/task/annotate.txt

**Input**: P—a set of semantically related words (positive examples), C—text corpus[10].
**Output**: PN—a balanced set of semantic relations similar to BLESS [3] with positive and negative examples for each target word.

1. Start with no negative examples: N = {}.
2. Calculate PMI-based noun similarity matrix **S** from the corpus C, where similarity between words $w_i$ and $w_j$:

$$s_{ij} = \log \frac{P(w_i, w_j)}{P(w_i)P(w_j)}$$

$$= \log \frac{\#w_i \text{ and } w_j \text{ coocurences in doc}}{\#\text{word coocurences in doc}} * \frac{\#\text{word occurences}}{\# w_i \text{ occurences}} * \frac{\#\text{word occurences}}{\# w_j \text{ occurences}}$$

3. Remove similarities greater than zero from **S**: $s_{ij} = max(0, s_{ij})$.
4. For each positive example $<w_i, w_j> \in$ P:
   - Candidates are relations from **S** with the source word: $\{<w_i, w_j> : w_i = source, s_{ij} > 0\}$.
   - Rank the candidates by target word frequency $freq(w_j)$:
   - Add two top relations $<w_i, w_k>$ and $<w_i, w_m>$ to negative examples N.
   - Remove all relations $<*, w_k>$ and $<*, w_m>$ from consideration: $s_{ij} = 0$, for all $i$ and $j \in \{k, m\}$.
5. Filter false negative relations with the help of human annotators. Each relation was annotated by at least two annotators. If at least one annotator indicates an error, remove this negative example from N.
6. The dataset PN is a union of positive and negative examples: $\{P \cup N\}$. Balance this dataset, so the number of positive and negative relations is equal for each source word.
7. Return PN.

The *Semantic Relation Classification* evaluation framework used here quantifies how well a system can distinguish related word pairs from unrelated ones. First, submitted word pairs are sorted by similarity. Second, we calculate the *average precision* metric [39]:

$$AveP = \frac{\sum P@r_r}{R}$$

Here $r$ is the rank of each relevant pair, $R$ is the total number of relevant pairs, and $P@r$ is the precision of the top-$r$ pairs. This metric is relevant as it takes ranking into account; it corresponds to the area under the *precision-recall curve* (see Fig. 1).

It is important to note that average precision of a random baseline for the semantic relation classification benchmarks *RT*, *AE* and *AE2* is 0.5 as these datasets are balanced (each word has 50% of related and 50% of unrelated candidates). Therefore, *RT*, *AE* and *AE2* scores should not be confused with semantic relation extraction evaluation, a task where the ratio of related and unrelated candidates and the average precision are close to 0.0.

---

[10] In our experiments we used Russian Wikipedia corpus to induce unrelated words.

**Table 3.** Structure of the semantic relation classification benchmarks (RT, AE, AE2)

| word1 | word2 | related |
|---|---|---|
| книга (book) | тетрадочка (notebook) | 1 |
| книга (book) | альманах (almanac) | 1 |
| книга (book) | сборничек (proceedings) | 1 |
| книга (book) | перекресток (crossroads) | 0 |
| книга (book) | марокко (marocco) | 0 |
| книга (book) | килограмм (kilogram) | 0 |

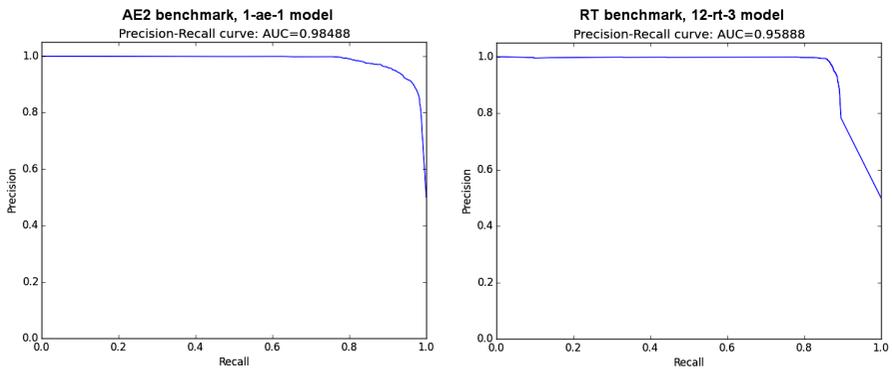

**Fig. 1.** Precision-recall curves of the best models on AE2 and RT datasets

### 3.3. Semantic Relation Classification of Associations (AE and AE2)

In the *AE* and *AE2* tasks, two words are considered similar if one is a cognitive (free) association of another. We used the results of two large-scale Russian *associative experiments* in order to build our training and test collections: the *Russian Associative Thesaurus*[11] (*AE*) and the Sociation.org (*AE2*). In an associative experiment, respondents were asked to provide a *reaction* to an input *stimulus*, e.g.:
- время, деньги, 14        (time, money, 14)
- россия, страна, 23        (russia, country, 23)
- рыба, жареная, 35        (fish, fried, 35)
- женщина, мужчина, 71   (woman, man, 77)
- песня, веселая, 33        (song, funny, 33)

The strength of an association is quantified by the number of respondents providing the same reaction. Associative thesauri typically contain a mix of synonyms, hyponyms, meronyms and other relations. Relations in such thesauri are often asymmetric.

---

[11]    http://it-claim.ru/Projects/ASIS/

To build the test sets we gathered 976 and 1,501 associations respectively from the *Russian Associative Thesaurus* and the *Sociation.org*. At this step, we used the target words with the highest association value between stimulus and reaction. Similarly to the *RT* dataset, we used only single-word nouns. Negative word pairs i.e. semantically unrelated words, were generated with the procedure described in the previous section. In the same fashion as the *RT*, we use average precision to measure the performance on the *AE* and *AE2* benchmark datasets.

## 4. Results and Discussion

Initially, 52 groups registered for the shared task, which shows high interest in the topic. A total of 19 teams finally submitted at least one model. These participants uploaded 105 runs (1 to 17 runs per team). A table with the evaluation results of all these submissions is available online[12]. To make the paper more readable, we present only abridged results here. First, we removed near duplicate submissions. Second, we kept only the best models of each participant. If one model was better than another with respect to all four benchmarks then the latter was dropped.

Participants used a wide range of approaches in order to tackle the shared task including:
- distributional models with context window and syntactic context: participants 3, 10, 11, 17;
- network-based measures that exploit the structure of a lexical graph: participants 2, 19;
- knowledge-based measures, including linguistic ontologies, Wiktionary and Wikipedia relations: participants 8, 12;
- measures based on lexico-syntactic patterns: participant 4;
- systems based on unsupervised neural networks, such as *CBOW* [26]: participants 1, 5, 7, 9, 13, 15, 16;
- supervised models: participants 1, 2, 5, 15.

These methods were applied to corpora of different sizes and genres (see Table 4), including Wikipedia, the Russian National Corpus (RNC), RuWaC, a news corpus, a web crawled corpus, a Twitter corpus, and three collections of books (Google N-Grams, Lib.ru, and Lib.rus.ec). Detailed descriptions of some submissions are available in the proceedings of the Dialogue 2015 conference[13].

Table 6 in the appendix presents the top 10 models according to the correlations with human judgements (*HJ*). The best results were obtained by the model *5-rt-3*[14], combining corpus-, dictionary-, and morpheme-based features. As one may observe, systems building upon *CBOW* and *skip-gram* models [26] trained on a big corpus yielded good results in this task. On the other hand, the classical distributional context window model *17-rt-1* also managed to find its place among the top results. Finally, the recent *GloVe* model *16-ae-1* also proved successful for the Russian language.

---

[12] http://russe.nlpub.ru/results

[13] http://dialog-21.ru/dialog2015, see the Dialogue Evaluation on semantic similarity.

[14] here *5-rt-3* is a submission identifier, where the first number (5) denotes the number of participant

**Table 4.** Russian corpora used by participants

| Corpus Name | Size, tokens |
|---|---:|
| Russian Wikipedia | 0.24 B |
| Russian National Corpus | 0.20 B |
| lib.rus.ec | 12.90 B |
| Russian Google N-grams | 67.14 B |
| ruWaC | 2.00 B |
| lib.ru | 0.62 B |

**Table 5.** 11 best models, sorted by the sum of scores. Each of the models is in top 5 of at least in one of the four benchmarks (HJ, RT, AE and AE2). Top 5 models are in bold font.

| Model ID | HJ | RT-AVEP | AE-AVEP | AE2-AVEP | Method Description |
|---|---|---|---|---|---|
| 5-ae-3 | 0.7071 | 0.9185 | 0.9550 | 0.9835 | Word2vec (skip-gram, window size 10, 300d vectors) on ruwac + lib.ru + ru-wiki, bigrams on the same corpus, synonym database, prefix dictionary, orthographic similarity |
| 5-rt-3 | **0.7625** | **0.9228** | 0.8887 | **0.9749** | Word2vec (skip-gram, window size 10, 300d vectors) on ruwac + lib.ru + ru-wiki, synonym database, prefix dictionary, orthographic similarity |
| 1-ae-1 | 0.6378 | **0.9201** | 0.9277 | 0.9849 | Desicion trees based on n-grams (Wikipedia titles and search queries), morphological features and Word2Vec |
| 15-rt-2 | 0.6537 | 0.9034 | 0.9123 | 0.9646 | Word2vec trained on 150G of texts from lib.rus.ec (skip-gram, 500d vectors, window size 5, 3 iteration, min cnt 5) |
| 16-ae-1 | 0.6395 | 0.8536 | 0.9493 | 0.9565 | GloVe (100d vectors) on RuWac (lemmatized, normalized) |
| 9-ae-9 | **0.7187** | 0.8839 | 0.8342 | 0.9517 | Word2vec CBOW with window size 5 on Russian National Corpus, augmented with skip-gram model with context window size 20 on news corpus |
| 17-rt-1 | **0.7029** | 0.8146 | 0.8945 | 0.9490 | Distributional vector-based model, window size 5, trained on RUWAC and NRC, plmi-weighting |
| 9-ae-6 | **0.7044** | 0.8625 | 0.8268 | 0.9649 | Word2vec CBOW model with context window size 10 trained on web corpus |
| 15-rt-1 | 0.6213 | 0.8472 | 0.9120 | 0.9669 | Word2vec trained on 150G of texts from lib.rus.ec (skip-gram, 100d vectors, window size 10, 1 iteration, min cnt 100) |
| 1-rt-3 | 0.4939 | **0.9209** | 0.8500 | **0.9723** | Logistic regression trained on synonyms, hyponyms and hypernyms on word2vec features with AUC maximization |
| 12-rt-3 | 0.4710 | **0.9589** | 0.5651 | 0.7756 | Applying knowledge extracted from Wikipedia and Wiktionary for computing semantic relatedness |

Results of the *RT* benchmark (synonyms and hypernyms) are summarized in Table 7 in the appendix. The first place belongs to a knowledge-based model that builds upon Wiktionary and Wikipedia. Otherwise, all other models at the top are either based on standard *word2vec* tools or on a hybrid model that relies on *word2vec* embeddings.

Tables 8 and 9 list models that were able to successfully capture cognitive associations. The supervised models *5-ae-3* and *1-ae-1* that rely on heterogeneous features, including those from *CBOW/skip-gram* models, showed excellent results on both *AE* and *AE2* benchmarks. Like in the other tasks, the *word2vec*, *GloVe* and distributional context window models show very prominent results.

Interestingly, the systems are able to better model associations (top 10 submissions of *AE2* ranging from 0.96 to 0.99) than hypernyms and synonyms (top 10 submissions ranging from 0.85 to 0.96) as exemplified in Tables 8 and 10. Therefore, semantics that is mined by the skip-gram model and other systems is very similar to that of cognitive associations.

Again, we must stress here that the average precision of semantic relation *classification* presented in Tables 5–9 should not be confused with the average precision of the semantic relation *extraction,* which is normally much lower. Our evaluation schema was designed to learn relative ranking of different systems.

Finally, Table 5 lists the 11 most successful systems overall, ranked by the sum of scores. Each model in this table is among the top 5 of at least one of the four benchmark datasets. The best models either rely on big corpora (ruWaC, Russian National Corpus, lib.rus.ec, etc.) or on huge databases of lexical semantic knowledge, such as Wiktionary. While classical distributional models estimated on a big corpus yield good results, they are challenged by more recent models such as *skip-gram, CBOW* and *GloVe*. Finally, supervised models show that it is helpful in this context to adopt an unsupervised model for a certain type of semantic relations (e.g. synonymy vs. association) and to combine heterogeneous features for other types.

## 5. Conclusions

The *RUSSE* shared task became the first systematic attempt to evaluate semantic similarity measures for the Russian language. The 19 participating teams prepared 105 submissions based on distributional, network, knowledge and neural network-based similarity measures. The systems were trained on a wide variety of corpora ranging from the Russian National Corpus to Google N-grams. Our main contribution is an open-source evaluation framework that relies on our four novel evaluation datasets. This evaluation methodology lets us identify the most practical approaches to Russian semantic similarity. While the best results in the shared task were obtained with complex methods that combine lexical, morphological, semantic, and orthographic features, surprisingly, the unsupervised skip-gram model trained a completely raw text corpus was able to deliver results in top 5 best submissions according to 3 of the 4 benchmarks. Overall, the experiments show that common approaches to semantic similarity for English, such as *CBOW* or distributional models, can be successfully applied to Russian.

Semantic similarity measures can be *global* and *contextual* [5]. While this research investigated global approaches for Russian language, in future research it would be interesting to investigate which contextual measures are most suited for languages with rich morphology and free word order, such as Russian.

## Acknowledgements

This research was partially supported by the ERC 2011 Starting Independent Research Grant n. 283554 (COMPOSES), German Research Foundation (DFG) under the project JOIN-T, and Digital Society Laboratory LLC. We thank Denis Egorov for providing the *Sociation.org* data; Yuri N. Philippovich, Andrey Philippovich and Galina Cherkasova for preparing the evaluation dataset based on the Russian Associative Thesaurus. We would like to thank all those who participated in the crowdsourced annotation process during the construction of the dataset of human judgement. We thank Higher School of Economics' students who annotated unrelated word pairs used in the evaluation materials. Finally, we thank Prof. Iryna Gurevych and Ilia Chetviorkin for their help with the design of the experimental tasks and user interface.

# Appendix 1.  The Best Submissions of the RUSSE Shared Task

**Table 6.** 10 best models according to the
HJ benchmark. Top 5 models are in bold font

| Model ID | HJ | Method Description |
|---|---|---|
| 5-rt-3 | **0.7625** | Word2vec (skip-gram, window size 10, 300d vectors) on ruwac + lib.ru + ru-wiki, synonym database, prefix dictionary, orthographic similarity |
| 9-ae-9 | **0.7187** | Word2vec CBOW with window size 5 on Russian National Corpus, augmented with skip-gram model with context window size 20 on news corpus |
| 5-ae-3 | **0.7071** | Word2vec (skip-gram, window size 10, 300d vectors) on ruwac + lib.ru + ru-wiki, bigrams on the same corpus, synonym database, prefix dictionary, orthographic similarity |
| 9-ae-6 | **0.7044** | Word2vec CBOW model with context window size 10 trained on web corpus |
| 17-rt-1 | **0.7029** | Distributional vector-based model, window size 5, trained on RUWAC and NRC, plmi-weighting |
| 15-rt-2 | 0.6537 | Word2vec trained on 150G of texts from lib.rus.ec (skip-gram, 500d vectors, window size 5, 3 iteration, min cnt 5) |
| 16-ae-1 | 0.6395 | GloVe (100d vectors) on RuWac (lemmatized, normalized) |
| 1-ae-1 | 0.6378 | Desicion trees based on n-grams (Wikipedia titles and search queries), morphological features and Word2Vec |
| 15-rt-1 | 0.6213 | Word2vec trained on 150G of texts from lib.rus.ec (skip-gram, 100d vectors, window size 10, 1 iteration, min cnt 100) |
| 1-rt-3 | 0.4939 | Logistic regression trained on synonyms, hyponyms and hypernyms on word2vec features with AUC maximization |
| 12-rt-3 | 0.4710 | Applying knowledge extracted from Wikipedia and Wiktionary for computing semantic relatedness |

**Table 7.** 10 best models according to the
RT benchmark. Top 5 models are in bold font

| Model ID | RT-AVEP | Method Description |
|---|---|---|
| 12-rt-3 | **0.9589** | Applying knowledge extracted from Wikipedia and Wiktionary for computing semantic relatedness |
| 5-rt-3 | **0.9228** | Word2vec (skip-gram, window size 10, 300d vectors) on ruwac + lib.ru + ru-wiki, synonym database, prefix dictionary, orthographic similarity |
| 1-rt-3 | **0.9209** | Logistic regression trained on synonyms, hyponyms and hypernyms on word2vec features with AUC maximization |
| 1-ae-1 | **0.9201** | Desicion trees based on n-grams (Wikipedia titles and search queries), morphological features and Word2Vec |
| 5-ae-3 | **0.9185** | Word2vec (skip-gram, window size 10, 300d vectors) on ruwac + lib.ru + ru-wiki, bigrams on the same corpus, synonym database, prefix dictionary, orthographic similarity |
| 15-rt-2 | 0.9034 | Word2vec trained on 150G of texts from lib.rus.ec (skip-gram, 500d vectors, window size 5, 3 iteration, min cnt 5) |
| 9-ae-9 | 0.8839 | Word2vec CBOW with window size 5 on Russian National Corpus, augmented with skip-gram model with context window size 20 on news corpus |
| 9-ae-6 | 0.8625 | Word2vec CBOW model with context window size 10 trained on web corpus |
| 16-ae-1 | 0.8536 | GloVe (100d vectors) on RuWac (lemmatized, normalized) |
| 15-rt-1 | 0.8472 | Word2vec trained on 150G of texts from lib.rus.ec (skip-gram, 100d vectors, window size 10, 1 iteration, min cnt 100) |
| 17-rt-1 | 0.8146 | Distributional vector-based model, window size 5, trained on RUWAC and NRC, plmi-weighting |

**Table 8.** 10 best models according to the
AE benchmark. Top 5 models are in bold font

| Model ID | AE-AVEP | Method Description |
|---|---|---|
| 5-ae-3 | **0.9550** | Word2vec (skip-gram, window size 10, 300d vectors) on ruwac + lib.ru + ru-wiki, big-rams on the same corpus, synonym database, prefix dictionary, orthographic similarity |
| 16-ae-1 | **0.9493** | GloVe (100d vectors) on RuWac (lemmatized, normalized) |
| 1-ae-1 | **0.9277** | Desicion trees based on n-grams (Wikipedia titles and search queries), morphological features and Word2Vec |
| 15-rt-2 | **0.9123** | Word2vec trained on 150G of texts from lib.rus.ec (skip-gram, 500d vectors, window size 5, 3 iteration, min cnt 5) |
| 15-rt-1 | **0.9120** | Word2vec trained on 150G of texts from lib.rus.ec (skip-gram, 100d vectors, window size 10, 1 iteration, min cnt 100) |
| 17-rt-1 | 0.8945 | Distributional vector-based model, window size 5, trained on RUWAC and NRC, plmi-weighting |
| 5-rt-3 | 0.8887 | Word2vec (skip-gram, window size 10, 300d vectors) on ruwac + lib.ru + ru-wiki, synonym database, prefix dictionary, orthographic similarity |
| 1-rt-3 | 0.8500 | Logistic regression trained on synonyms, hyponyms and hypernyms on word2vec features with AUC maximization |
| 9-ae-9 | 0.8342 | Word2vec CBOW with window size 5 on Russian National Corpus, augmented with skip-gram model with context window size 20 on news corpus |
| 9-ae-6 | 0.8268 | Word2vec CBOW model with context window size 10 trained on web corpus |
| 12-rt-3 | 0.5651 | Applying knowledge extracted from Wikipedia and Wiktionary for computing semantic relatedness |

**Table 9.** 10 best models according to the
AE2 benchmark. Top 5 models are in bold font

| Model ID | AE2-AVEP | Method Description |
|---|---|---|
| 1-ae-1 | **0.9849** | Desicion trees based on n-grams (Wikipedia titles and search queries), morphological features and Word2Vec |
| 5-ae-3 | **0.9835** | Word2vec (skip-gram, window size 10, 300d vectors) on ruwac + lib.ru + ru-wiki, big-rams on the same corpus, synonym database, prefix dictionary, orthographic similarity |
| 5-rt-3 | **0.9749** | Word2vec (skip-gram, window size 10, 300d vectors) on ruwac + lib.ru + ru-wiki, synonym database, prefix dictionary, orthographic similarity |
| 1-rt-3 | **0.9723** | Logistic regression trained on synonyms, hyponyms and hypernyms on word2vec features with AUC maximization |
| 15-rt-1 | **0.9669** | Word2vec trained on 150G of texts from lib.rus.ec (skip-gram, 100d vectors, window size 10, 1 iteration, min cnt 100) |
| 9-ae-6 | 0.9649 | Word2vec CBOW model with context window size 10 trained on web corpus |
| 15-rt-2 | 0.9646 | Word2vec trained on 150G of texts from lib.rus.ec (skip-gram, 500d vectors, window size 5, 3 iteration, min cnt 5) |
| 16-ae-1 | 0.9565 | GloVe (100d vectors) on RuWac (lemmatized, normalized) |
| 9-ae-9 | 0.9517 | Word2vec CBOW with window size 5 on Russian National Corpus, augmented with skip-gram model with context window size 20 on news corpus |
| 17-rt-1 | 0.9490 | Distributional vector-based model, window size 5, trained on RUWAC and NRC, plmi-weighting |
| 12-rt-3 | 0.7756 | Applying knowledge extracted from Wikipedia and Wiktionary for computing semantic relatedness |